%% file: main.tex
\documentclass{article}
\usepackage[utf8]{inputenc}
\usepackage[graphicx]{realboxes}
\usepackage{booktabs}
\usepackage[export]{adjustbox}
\usepackage[hyphens]{url}
\usepackage[hidelinks]{hyperref}
\usepackage{amsmath}
\usepackage{mathtools}
\usepackage{multirow}
\usepackage{tabularx}
\usepackage{caption}
\usepackage[table,xcdraw]{xcolor}
\usepackage{tabularx}
\usepackage{mathpazo}

\usepackage{setspace}
\usepackage{authblk}
\usepackage{graphicx}
\usepackage[margin=0.75in]{geometry}
\usepackage{svg}

\usepackage{hyperref}
\usepackage{xfp}
\newcommand\SupplementaryMaterials{%
  \xdef\presupfigures{\arabic{figure}}
  \xdef\presupsections{\arabic{section}}
  \xdef\presuptables{\arabic{table}}
  \renewcommand\thefigure{S\fpeval{\arabic{figure}-\presupfigures}}
  \renewcommand\thesection{S\fpeval{\arabic{section}-\presupsections}}
  \renewcommand\thetable{S\fpeval{\arabic{table}-\presuptables}}
}

\DeclareUnicodeCharacter{00B1}{\ensuremath{\pm}}
\DeclareUnicodeCharacter{271D}{\ensuremath{\dagger}}

\title{Adversarial confound regression and uncertainty measurements to classify heterogeneous clinical MRI in Mass General Brigham}

\author[1,2,*]{Matthew Leming}
\author[2,3]{Sudeshna Das}
\author[1,2,4,$\dagger$]{Hyungsoon Im}
\date{\today}
\affil[1]{Center for Systems Biology, Massachusetts General Hospital, Boston, MA, USA}
\affil[2]{Massachusetts Alzheimer's Disease Research Center, Charlestown, MA, USA}
\affil[3]{Department of Neurology, Massachusetts General Hospital, Boston, MA, USA}
\affil[4]{Department of Radiology, Massachusetts General Hospital, Boston, MA, USA}
\affil[*]{Corresponding author: Matthew Leming, mleming@mgh.harvard.edu}
\affil[$\dagger$]{Corresponding author: Hyungsoon Im, im.hyungsoon@mgh.harvard.edu}

\begin{document}
\captionsetup[figure]{labelfont={bf},name={Fig.},labelsep=period}
\captionsetup[table]{labelfont={bf},name={Tab.},labelsep=period}
\maketitle
\begin{abstract}

Automated disease detection in neuroimaging holds promise to improve the diagnostic ability of radiologists, but routinely collected clinical data frequently contains technical and demographic confounding factors that cause data to both differ between sites and be systematically associated with the disease of interest, thus negatively affecting the robustness of diagnostic models. There is a critical need for diagnostic deep learning models that can train on such imbalanced datasets without being influenced by these confounds. In this work, we introduce a novel deep learning architecture, MUCRAN (Multi-Confound Regression Adversarial Network), to train a deep learning model on clinical brain MRI while regressing demographic and technical confounding factors. We trained MUCRAN using 17,076 clinical T1 Axial brain MRIs collected from Massachusetts General Hospital before 2019 and demonstrated that MUCRAN could successfully regress major confounding factors in the vast clinical data. We also applied a method for quantifying uncertainty across an ensemble of these models to automatically exclude out-of-distribution data in the AD detection. By combining MUCRAN and the uncertainty quantification method, we showed consistent and significant increases in the AD detection accuracy for newly collected MGH data (post-2019) and for data from other hospitals. MUCRAN offers a generalizable approach for heterogenous clinical data for deep-learning-based automatic disease detection.

\end{abstract}

\begin{center}
\end{center}

\section{Introduction}

The use of AI to diagnose diseases from brain MRI holds promise to automate, standardize, and apply the diagnostic process at scale. Because clinical MRI are collected routinely, they amass into large databases that can be utilized to train such AI algorithms. Deep learning in particular has shown success in detecting multiple diseases in high-quality brain MRI data collected in a controlled research setting \cite{Falkai2018,Wen2020}. However, extending this diagnostic technology to deep learning in clinical settings is hampered by practical challenges \cite{Gollub2021}. Compared to images collected in research settings, clinical imaging data is often lower in quality and more diverse in technical variables, labeling, diseases, and patient populations. Furthermore, machines and clinical techniques used to acquire data differ between settings. Hence, a deep-learning model trained on clinical data from one hospital will not necessarily generalize to data from another. Imbalances in technical and demographic variables, if not carefully accounted for, could lead a deep-learning model to fit to confounding factors rather than true biomarkers.

Implementing deep-learning models across hospitals requires a large degree of model robustness and ability to scale \cite{Elemento2021}. This is a form of the out-of-distribution data problem \cite{Lee2018} that is endemic in the field of deep learning. Data privacy concerns unique to healthcare make this problem especially difficult to overcome in diagnostic models, as the pooling of data in healthcare is yet infeasible from a policy standpoint. Sophisticated training strategies, such as federated learning \cite{Dayan2021}, in which models are trained internally in healthcare systems and subsequently released, address this robustness problem to an extent. Even with such sophisticated training strategies, model robustness via diversification of data would not necessarily aid cases in which a given confound is systematically associated with a particular disease (e.g., age and degenerative diseases). 

Several methods have been proposed to overcome the problems associated with confounded data in neuroimaging. We have previously used a data matching scheme \cite{Leming2021} to regress confounding factors in clinical MRI data by curating confounder-free, matched datasets for three-stage AD classification \cite{Leming2022}. Data matching, however, causes the training set size to decrease as more confounding factors are included in the model. Other methods, such as ComBat \cite{Johnson2007,Yu2018} and Linear mixed-effects models \cite{Bolker2009,EspinPerez2018}, regress confounding effects directly but only work on scalar features extracted from images and so are highly dependent on the chosen methods of feature extraction. Deep-learning-based regression methods, by contrast, can be applied directly to input images \cite{Zhao2020,Kimmel2021}. Zhao et al. \cite{Zhao2020} proposed a confounder-free neural network trained using three optimizers in an adversarial scheme, demonstrating its effectiveness using scalar demographic confounding factors, but these methods, as proposed, can only be applied to one confounding factor at a time.

Another practical issue in deep-learning-based disease detection is out-of-distribution samples. In prospective applications, even a robust deep learning model could still be vulnerable to new or unknown confounding factors or new data that fall outside the purview of the original training data (e.g., images acquired with a new MRI scanner). Uncertainty estimation to detect out-of-distribution samples is a recent area of interest in deep learning \cite{Lee2018,Shad2021}, which has seen the development of sophisticated means of measuring uncertainty \cite{Gal2016,Liu2020,Gawlikowski2021}). Thus, for a real-world diagnostic system, it is necessary to design a deep learning model that can classify by disease without the influence of both confounds and out-of-distribution samples.

In this work, we developed the Multi-Confound Regression Adversarial Network (MUCRAN): an adversarial process on a specialized scheduler to train a model on a vast clinical dataset while regressing multiple confounding factors. We trained MUCRAN on 17,076 clinical brain MRIs and successfully regressed 11 demographic and technical confounding factors that could  adversely affect AD detection. We also implemented an uncertainty quantification method by training multiple independent models in an ensemble and making a consensus decision. Along with MUCRAN, the integrated approach showed significantly improved AD detection accuracies for recently collected MGH data and data from other hospitals.

\section{Results}


\subsection{Regressing confounding factors using MUCRAN}

We designed MUCRAN to regress multiple confounding factors in the vast clinical data. MUCRAN takes a 3D image as input and outputs a one-hot vector for a disease label, as well as each confound included in the training (Fig \ref{fig:gan_pipeline_fig}A). In its training, MUCRAN is incentivized to classify an output label (AD or non-AD) without the use of confounding factors (see Methods for a detailed description). MUCRAN consists of an encoder and regressor, similar, respectively, to a generator and discriminator in conventional generative adversarial neural networks (GANs) \cite{Goodfellow2014}. The encoder translates the input to an intermediary feature representation, and the regressor translates these features to predictions of the disease label and confounding factors (e.g., age, sex, image modality), represented by a 2-D array of one-hot vectors. In the training scheme, back-propagation is applied to the regressor and encoder alternately using different output arrays (Fig \ref{fig:gan_pipeline_fig}B); the regressor is trained using true output confound/label encodings, while the encoder is trained using the true label encoding, but confound encodings that are all set to the same value. Thus, the encoder is incentivized to output an intermediary feature representation from which a true label, but no confounding factors, can be derived.

\input{figures/gan_pipeline_fig}

For uncertainty quantification, we employed a consensus approach to separate out-of-distribution data (unlike the training set) from in-distribution data (similar to the training set). We trained ten models independently in an ensemble; for each test, ten disease predictions were output (Fig \ref{fig:gan_pipeline_fig}C). After a softmax layer, the sum of each of these predictions were normalized to 1. These ten predictions were averaged across the ensemble, which fell in the range between 0.5 and 1.0. As individual measurements are averaged across an ensemble of models, averaged in-distribution measurements tend towards either 0.0 or 1.0, indicating that all models agree on a classification, while averaged out-of-distribution measurements tend towards 0.5, indicating that the outputs are essentially random. Thus, in-distribution measurements can be isolated by removing outputs with an average that falls below a certain threshold value. For comparison, we show the differences in model accuracy on both the entire test set (threshold = 0.5) and for just the "in-distribution" portion (threshold = 0.9).



We trained MUCRAN with T1 Axial MRI data from Massachusetts General Hospital (MGH, n = 17,076) collected between 1995 and 2018 (Pre-2019, Fig \ref{fig:dataset_description}). We kept the MGH data collected between 2019-2021 (Post-2019, n = 1,497) out of the training and used it as an internal test set. This is to test the performance of the trained model for newly acquired clinical image data.

\input{figures/dataset_description}


We next investigated the regression of multiple confounding factors in MUCRAN. Of the 141 variables present in the dataset, we selected eleven confounding factors to be regressed from the model: age, employment status, ethnic group, marital status, patient class (e.g., inpatient or outpatient), religion, sex, specific absorption rate, imaging frequency, pixel bandwidth, and repetition time. Several criteria were considered in the selection of these confounding factors, including (1) their variance and distribution across T1 Axial MRIs (single-valued confounding factors were not included); (2) a low number of categorical choices, which made it practically encodable; (3) their presence across a large amount of data; (4) their theoretical relevance to both site differences and AD; and (5) their likelihood of being predicted from MRI, with confounding factors that were likely to be predicted from MRI (i.e., age, sex) and unlikely (i.e., religion) both included as a means of comparison. 

Figure \ref{fig:confounds_diagram} shows model performance for predicting demographic and technical confounding factors by MRIs. In the confounded model, in which confounding factors were predicted directly, some confounding factors, such as age, sex, imaging frequency, pixel bandwidth, and specific absorption rate, could be predicted very effectively, while others, such as ethnic group, marital status, and religion, could not be predicted, as expected. In MUCRAN, however, it shows that the the areas under the receiver operating characteristic curves (AUROCs) for all confounding factors are within the 10\% margin from 0.5. This indicates that the regressed model fails to predict confounding factors using MRI data. It means MUCRAN largely achieved its goal of making a set of intermediary features from which confounds could not be predicted, and thus the adverse effects from confounding factors are minimized in AD detection.

\input{figures/confounds_diagram}

\subsection{AD classification}


We next applied MUCRAN for AD detection. To test our model, we constructed both internal and external test sets. MUCRAN was trained on a dataset from MGH collected before 2019, and our internal test set consisted of data from MGH data collected after 2019 (MGH Post-2019, Fig \ref{fig:dataset_description}). We also constructed two external test sets, consisting of data from Brigham and Women's Hospital and data imported from outside hospital systems. This is to test, first, how the regressed model performs in MRI-based AD detection prospectively in a given data set and, second, how much our uncertainty approach could improve the detection accuracy for data collected in different settings.

In each of our test sets, there is a significant  difference in age distributions between AD and non-AD groups (Fig \ref{fig:dataset_description}). We previously showed that unmatched clinical data sets could lead to artificial gains in model performance for AD classification\cite{Leming2022}. Therefore, we samples each of our test sets for age-matched datapoints of equal size between AD and non-AD groups. Additionally, because age is one of the most significant risk factors for AD, we only included patients with ages above 55 (a higher age threshold produces much smaller data sets, which is unsuitable for robust accuracy testing).

For a comparative analysis, the performance of MUCRAN was compared with two other models - "baseline" and "confounded." The baseline model refers to a model for which only the Alzheimer's label was predicted. The confounded model refers to a model for which the Alzheimer's label, as well as the 11 confounds, were predicted directly. Only MUCRAN attempted to regress the confounding factors.

Figure \ref{fig:ad_age_acc_comp}A shows the AD classification results for the internal test set (MGH post-2019). First, without uncertainty thresholding, all three models showed poor classification accuracies below 70\%. The accuracies were improved more than 10\% when an uncertainty threshold of 0.9 was applied to isolate in-distribution data. For the in-distribution data, MUCRAN outperformed two other models by a margin of 12\% and achieved the accuracy of 85\%. The results show that both regressing confounding factors and excluding out-of-distribution data are critical to achieve robust classification accuracy for clinical MRI data. 

\input{figures/ad_age_acc_comp.tex}

Next, we applied the three models for data collected from Brigham and Women's Hospital (BWH) and other hospitals (Figures \ref{fig:ad_age_acc_comp}B and C). Similar to the post-2019 MGH data, MUCRAN outperformed two other models by more than 10\% for the in-distribution data. The accuracy was increased from 72\% in the baseline model to 90\% in MUCRAN for data from BWH; it was increased from 71\% to 81\% for data from other hospitals. Thus, trained only on pre-2019 MGH data and using the uncertainty thresholding, our models maintain a classification accuracy over 80\% for data collected in different settings.

Table \ref{tab:results_alz} summarizes the AD classification results and sample sizes for all five test sets as well as in the combined set, both for the age-matched sample shown in Figure \ref{fig:ad_age_acc_comp} and the entire test set shown in Figure \ref{fig:dataset_description}. Across the pooled test sets, MUCRAN outperformed the comparative models, both in the age-matched sample (75.3\% versus 67.4\% and 61.8\%) and the whole test set (89.1\% versus 84.7\% and 83.9\%). Between these five test sets, there were variations in classification accuracy, which can be explained by a number of factors that, in themselves, would lead to both increases and decreases in accuracy. For instance, the MGH post-2019 data contains many sites that are the same as the MGH pre-2019 dataset, a factor that aids its accuracy in the confounded model, which overfits to individual sites (89.5\% for MGH post-2019 confounded versus 87.6\% for BWH post-2019 confounded) but not MUCRAN (92.5\% for MGH post-2019 confounded MUCRAN versus 93.5\% for BWH post-2019 MUCRAN). The post-2019 datasets likely contain more high-quality images from modern scanners that would aid classification accuracy, while the pre-2019 sets have historical data that are both low-quality and poorly organized, leading to mismatched labels (89.2\% BWH pre-2019 MUCRAN versus 93.5\% BWH post-2019 MUCRAN); this latter point, however, is likely of greater concern for the Brigham and Women's pre-2019 dataset, since data that clinicians took the trouble to import from outside hospital systems would have received more scrutiny in terms of its diagnostic usefulness (89.2\% for BWH pre-2019 versus 92.1\% for Others, pre-2019). In short, there are a myriad of reasons that may explain why certain training methods out- or under-performed others for different time-based and site-based splits of the test set. On average, however, for this task, MUCRAN outperformed the confounded and baseline models, and, given the vast differences in classification accuracy on the older age-matched sample, this is likely due to imbalances in age in each of the test sets being slightly different from the imbalances in age in the training set -- a situation for which MUCRAN is specifically designed to account for.

\input{tables/results_alz_55}

In terms of sample sizes, MUCRAN's in-distribution dataset was consistently smaller than those of the two comparison models (857 for MUCRAN, 1317 for Baseline, and 1655 for confounded). In the pooled test set, MUCRAN's in-distribution test set was 5783/20857 while the baselines' was 9469/20857, a 39\% decrease. This is likely indicative that the MUCRAN ensemble did not so much generalize to more data as be skeptical about making "bad hires" for its in-distribution pool.

\subsection{Sex}

Table \ref{tab:results_sex} show the results for sex classification. In this test, MUCRAN showed a much higher classification accuracy than AD classification. This is evidence especially of the effectiveness of uncertainty thresholding in instances where the label is consistent and has a definite biological basis, as this led from an 85.7\% accuracy to a 96.1\% accuracy, the highest of any of the selected tasks. It is very likely that labels in the other two tasks were, in many cases, imperfect and incomplete. Unlike biological sex, which is consistently recorded in the electronic health record, many factors complicate the Alzheimer's and localized anomaly label. ICD codes may be inconsistently recorded, and medication history is only an imperfect indication of Alzheimer's; head trauma, even if recorded, may not leave a biological mark evident in a brain MRI. On the other hand, sex is nearly always present in an electronic health record and there is a definite biological basis for sex classification in structural brain MRI. This task thus offers an insight on what can be determined in clinical MRI in cases where the label is consistent and known biomarkers are present.

\input{tables/results_sex}

\section{Discussion}

This work focused on designing a model disincentivized from incorporating confounding factors in its classification decision and developing strategies to isolate in-distribution parts of a given test set. Unlike research settings where large amounts of data are curated for certain deep learning tasks, clinical data represent highly heterogeneous, often poorly-organized sets that contain many different confounding factors and labels that may be only tenuously associated with the underlying ground-truth. With MUCRAN, we addressed the issue of confounding factors by designing adversarial networks to regress and minimize their influence in the model performance, and with our uncertainty thresholding method, we showed a way of honing in on subsets of a test set that a given model is more likely to be able to classify correctly. This approach is uniquely suited to clinical imaging data.




Table \ref{tab:methods_comparison_table} shows a comparison of MUCRAN with other recently proposed variable regression methods. Methods to control for confounds occur in many different ways, including in data collection methods, ranging from cross- sectional study design \cite{Cook2002} to the use of special devices to control for head motion in MRI, as well as post-hoc methods designed to control for specific confounds, such as computational methods in registration and motion regression. Table \ref{tab:methods_comparison_table} specifically covers those methods that are (1) generalizable to any given confound; (2) may be applied post-hoc; and (3) assume that the given confound is recorded. The most similar method, proposed in Zhao 2020 et al, is less effective in regressing multiple confounding factors, since it uses a third loss function for its confound regression; early tests with this showed that this frequently led to mode collapse, a common problem when training adversarial networks. MUCRAN, in contrast, only uses two loss functions and is structured more analogously to GANs, which, as stated in the Methods, allows it to take advantage of most of the modern developments in GAN training.

\input{tables/methods_comparison_table}

Our results showed that MUCRAN was most effective in regressing the effects of multiple confounding factors systematically associated with the label being classified for. In this case, the clearest association is age and Alzheimer's, a degenerative disease. Age-matched results showed that MUCRAN outperformed the confounded and baseline models by a substantial margin across all five test sets, and by an average of 13.5\% higher than the confounded model and and 7.9\% higher than the baseline (Table \ref{tab:results_alz}). Even so, it is likely that many of the confounding factors introduced by site differences were largely addressed by the diversity of data in the MGH pre-2019 training set, since both the confounded and baseline models also performed relatively well across sites on the five test sets.

Our uncertainty thresholding method removed out-of-distribution or intermediary cases from the test set, as quantified by the average value of a given classification task across an ensemble of models. Test set accuracy rose sharply on those remaining in-distribution datapoints, across the three types of models tested, making a strong case for use of this diagnostic technology in a clinical setting. The accuracy of these methods approached that typically achieved in deep learning studies in Alzheimer's on research-grade MRI datasets, such as ADNI and AIBL \cite{Wen2020}, which usually achieve between 85 and 90 percent accuracy, depending on the model hyperparameters and inclusion/exclusion criteria of the particular study used. More importantly, we showed that the AD classification accuracies remain over 80\% for other hospital data while the confounded model showed worse performances (56-67\%) in other hospitals compared to MGH data (72\%). 



The deep learning task presented was challenging from a bioinformatics perspective. The inference of Alzheimer's disease in clinical records is difficult because they are often improperly labeled; ICD codes may be incomplete, and data recording practices vary with databases, clinics, time periods, and medical practitioners. While it is an imperfect marker, the use of medications as a label marker was advantageous because (1) the four medications used in this study are used to treat Alzheimer's in its different stages (though Memantine is sometimes used in younger age groups \cite{Hosenbocus2013}) and (2) prescriptions are consistently recorded in the electronic health record. We included, as well, instances in which an ICD indicating Alzheimer's was recorded but not a medication.

While most deep learning tasks present one straightforward metric to improve — accuracy — the current study presents two: both accuracy and test set inclusion. Overall, the regressed model, after thresholding, made predictions on an in-distribution dataset consisting of 5873/20857 datapoints, or 28.2\% of the data (Table \ref{tab:results_alz}), at 89.1\% accuracy (though, considering only post-2019 data, which omits much of the lower-quality legacy MRIs in the dataset, this percentage rises to above 90\% in all cases). On the one hand, accurate predictions on 28.2\% of MRIs at a given hospital for routinely-collected MRI would be a useful metric for radiologists to consider; on the other hand, this still omits over two-thirds of all data. Even so, unlike a system that measures every datapoint considered, whether or not it is in- or out-of-distribution, ensembles of MUCRANs, by offering fewer unreliable predictions, present a system that can be implemented and scaled in the real world.

In conclusion, we present deep learning methods that are able to generalize across dates and hospitals for diagnosing Alzheimer's disease in complex clinical MRI. We also show a means of separating out-of-distribution data that cannot be effectively assessed by our models from in-distribution data that can, providing a scalable AI system that can be deployed in new environments.

\section{Materials and Methods}
\subsection{Data}

We used an extremely large amount of clinical data from a diverse array of MRI scanners, meaningfully split into training and tests sets, separated by hospital and time period, to test for cross-site and cross-time generalizability. Descriptions of the full dataset can be seen in Figure \ref{fig:dataset_description}. MRI data were requested from the Mass General Brigham Research Patient Data Registry (RPDR), and additional variables were augmented from the Enterprise Data Warehouse (EDW), which stored additional patient metadata but no images. Data were separated into three sets: MGH, BWH, and Other (i.e., miscellaneous data which were imported by patients from outside hospital systems). These were further subdivided by time period (pre- and post-2019 data). Pre-2019 MGH data were used for training and the rest were used for testing.

The presence of Alzheimer's disease was assumed by analyzing patient medication records, in particular the presence of Galantamine, Rivastigmine, Donepezil, or Memantine, or an ICD 10 code of G35. While medication does not constitute a precise diagnosis of Alzheimer's, it was more consistently recorded across the database than ICD codes. Patients with ICD-10 codes for a malignant neoplasm of the brain (C71.1, C71.9, C79.31), cerebral infarction (I63.9), neoplasm of unspecified behavior of brain, (D49.6), benign neoplasm of cerebral meninges (D32.0), and previous head trauma (S00 - S09) were designated into a third group, indicating patients with localized anomalies in the brain that could likely be seen by a human interactor. These instances were excluded. Finally, in order to offer a baseline, non-disease-related classification task, for which labels are consistently present, we also classify by biological sex within the control group.

\subsection{Preprocessing}

Due to the size and variability of the dataset, preprocessing was limited to translating images from DICOM to Nifti (dcm2niix), reorienting them to a standard space (fslreorient2std), and resizing them to a standard $96\times 96\times 96$ dimension. Diversity of the dataset led to a number of preprocessing errors, and so a number of criterion were made to include data, notably the size of the file (i.e., files that were too small were excluded from consideration -- these technical "exclusions" are not included in Figure \ref{fig:dataset_description}.) Dataset size and variability prevented the application of conventional MRI preprocessing methods, such as registration to a template.


\subsection{ML Model and training}

As shown in Figure \ref{fig:gan_pipeline_fig}B, MUCRAN is trained adversarially to classify by an output label while regressing the selected confounds. The model's structure is similar to generative adversarial networks (GANs) \cite{Goodfellow2014}. Briefly, GANs are incentivized to generate photorealistic images using a generator, which outputs the images, and a discriminator, which is trained to discriminate between real images and the generator's outputs; by training both in an adversarial process, the generator eventually outputs images that the discriminator is unable to distinguish. In MUCRAN, the "generator" is an encoder that translates input images to an intermediary feature representation (the "image"), while the "discriminator" is a regressor that translates these features from the intermediary feature representation to predictions of both the label and a number of confounding factors (i.e., age, sex, image modality -- the "real/fake" prediction). The regressor is trained using true output label/confound encodings, while the encoder is trained using the true label but confounds that are all set to the same value. In this way, the encoder is incentivized to output an intermediary feature representation from which a true label, but no confounding factors, can be derived.

Put formally, suppose we have input data, $x$, a label, $y$, with possible values ${y_1,y_2,...y_N}$, and a number of confounds, ${c^1,c^2,...c^K}$, each with possible values ${c^i_1,c^i_2,...c^i_N}$. The encoder, $E$, outputs intermediary features, $E(x) = F$, while the regressor, $R$, outputs a vector that combines the label and confounds, such that $R(F) = p( [y, c^1,c^2,...c^K])$. The loss function of the encoder is:

\begin{equation}
Loss(E(x)) = \begin{bmatrix}W \\ 1 \\ 1 \\ \vdotswithin{\ldots} \\ 1 \end{bmatrix}\frac{-1}{N} \cdot \sum_{i=1}^{N} \left(\begin{bmatrix}y_i \\ G(i) \\ G(i) \\ \vdotswithin{\ldots} \\ G(i) \end{bmatrix} \cdot log(R(E(x)) +  1 - \begin{bmatrix}y_i \\ G(i) \\ G(i) \\ \vdotswithin{\ldots} \\ G(i) \end{bmatrix} \cdot log(1 - R(E(x))))\right)
\label{eq:eq1}
\end{equation}

Where \[
    G(x)= 
\begin{cases}
    1,& \text{if } x = 1\\
    0,& \text{if } x\neq1
\end{cases}
\]

While the loss function of the regressor is:

\begin{equation}
Loss(R(x)) = \begin{bmatrix}W \\ 1 \\ 1 \\ \vdotswithin{\ldots} \\ 1 \end{bmatrix}\frac{-1}{N} \cdot \sum_{i=1}^{N} \left(\begin{bmatrix}y_i \\ c_i^1 \\ c_i^2 \\ \vdotswithin{\ldots} \\ c_i^K \end{bmatrix} \cdot log(R(E(x)) +  1 - \begin{bmatrix}y_i \\ c_i^1 \\ c_i^2 \\ \vdotswithin{\ldots} \\ c_i^K \end{bmatrix} \cdot log(1 - R(E(x))))\right)
\label{eq:eq2}
\end{equation}

These are both modified binary crossentropy loss functions. To ensure its convergence, the $y$ label is given an additional weighting factor, $W$ (in practice, this is set to 6).

The similarity of these models to GANs allowed us to draw on the wide body of research and conventions used to train them \cite{Chintala2016,Radford2016,Salimans2016}. The encoder was trained using an Adam optimizer, while the regressor was trained using an SGD optimizer. Sparse gradients, such as ReLU and max pooling, were avoided in the construction of the networks. The use of an adversarial system placed certain limitations on which layers could be used; for instance, pooling layers were removed in favor of strided convolutions, and batch normalization layers had to be avoided at the output of the encoder and input of the regressor \cite{Radford2016}. Extensive testing was performed on more sophisticated network architectures. However, with the adversarial training process, more complex CNNs, such as ResNet, DenseNet, and InceptionNetV3, failed to converge, or simply performed worse than the simple layered CNN used in this work (Fig \ref{fig:gan_pipeline_fig}A).

To add a comparison in our analysis, three total classes of models were trained, which are referred to as "MUCRAN", "baseline", and "confounded". These models are all the same structure, but were trained differently by modifying the loss function of the encoder and regressor. MUCRAN was trained using the loss functions as presented above (i.e., Equation \ref{eq:eq1} for the encoder and equation \ref{eq:eq2} for the regressor); the confounded versions were trained using equation \ref{eq:eq2} for both the encoder and regressor; and the baseline model was trained using equation \ref{eq:eq1} for both (effectively making it a standard, AlexNet-style CNN that only classifies by the given label). In effect, the confounded model predicts both the label and confounds, the baseline model predicts only the label, and MUCRAN, the only model trained using a truly adversarial process, predicts the label while regressing the confounds.

\subsection{Batch scheduler}

Training was not performed in epochs over the whole training set, as is typically the case. Rather, it was implemented using a scheduler that maintained equal ratios in any given batch between different label values. The scheduler constructed individual batches, half with equal ratios between classes of the label, and the other half with equal ratios of iterative confounding factors (thus, for Alzheimer's classification, 50\% of one batch would be composed of data that is half Alzheimer's and half control, while the other 50\% would have equal ratios of male and female; in the next batch 50\% would be half Alzheimer's and half control and the other 50\% equal distributions of age; and so on). 48 MRIs of these batches were then loaded into main memory at a time and trained for five iterations in random order. After a batch had been completely loaded and discarded from main memory, a new one would then be sampled. This process was repeated until 33,000 datapoints had been loaded and trained on.

To make the MUCRAN, baseline, and confounded model classes more comparable, they were trained side-by-side, with data fed in the same order.


A test set may consist of in-distribution similar to the training set and out-of-distribution data that is unlike the training set. To evaluate our test sets, a consensus approach was applied that separated out-of-distribution data from in-distribution data. Ten models were trained independently in an ensemble; for each point evaluated in the test set, ten predictions for each label were output. After a softmax layer, the sum of each of these predictions were normalized to 1. These ten predictions were averaged across the ensemble, and the averaged prediction with the higher value was taken as the final prediction for that datapoint. This averaged prediction fell in the range between 0.5 and 1.0 (see Figure \ref{fig:gan_pipeline_fig}C for an illustration of this thresholding). As individual measurements are averaged across an ensemble of models, averaged in-distribution measurements tend towards either 0.0 or 1.0, indicating that all models agree on a classification, while averaged out-of-distribution measurements tend towards 0.5, indicating that the outputs are essentially random. Thus, after averaging predictions using an ensemble of models, in-distribution measurements can be isolated by removing outputs with an average that falls below a certain threshold value (i.e., varies across ensembles). In the reported results for all tests, we show the differences in model accuracy on both the entire test set (threshold = 0.5) and for just the "in-distribution" portion (threshold = 0.9).

\section*{Data availability}

The clinical data utilized in this study is not publicly available because it contains confidential information that may compromise patient privacy as well as the ethical or regulatory policies of our institution. A part of data will be made available on reasonable request by contacting the corresponding author (H.I. im.hyungsoon@mgh.havard.edu) upon the approval from institutional review boards.  

\section*{Code availability}

Our code used in this study is publicly available at \url{https://github.com/mleming/MUCRAN}.

\section*{Acknowledgements}
This study was funded by U.S. NIH grant P30AG062421, R01GM138778, and the Technology Innovation Program (20009571) funded by the Ministry of Trade, Industry and Energy, Republic of Korea, managed through a subcontract to Massachusetts General Hospital 

\bibliography{references}
\newpage
\SupplementaryMaterials

\section*{Supplementary Information}




\subsection*{Training set sites}

\input{tables/sites}

\end{document}

%% file: figures/gan_pipeline_fig.tex
\begin{figure}[!h]
\centering
\includegraphics[width=0.80\textwidth]{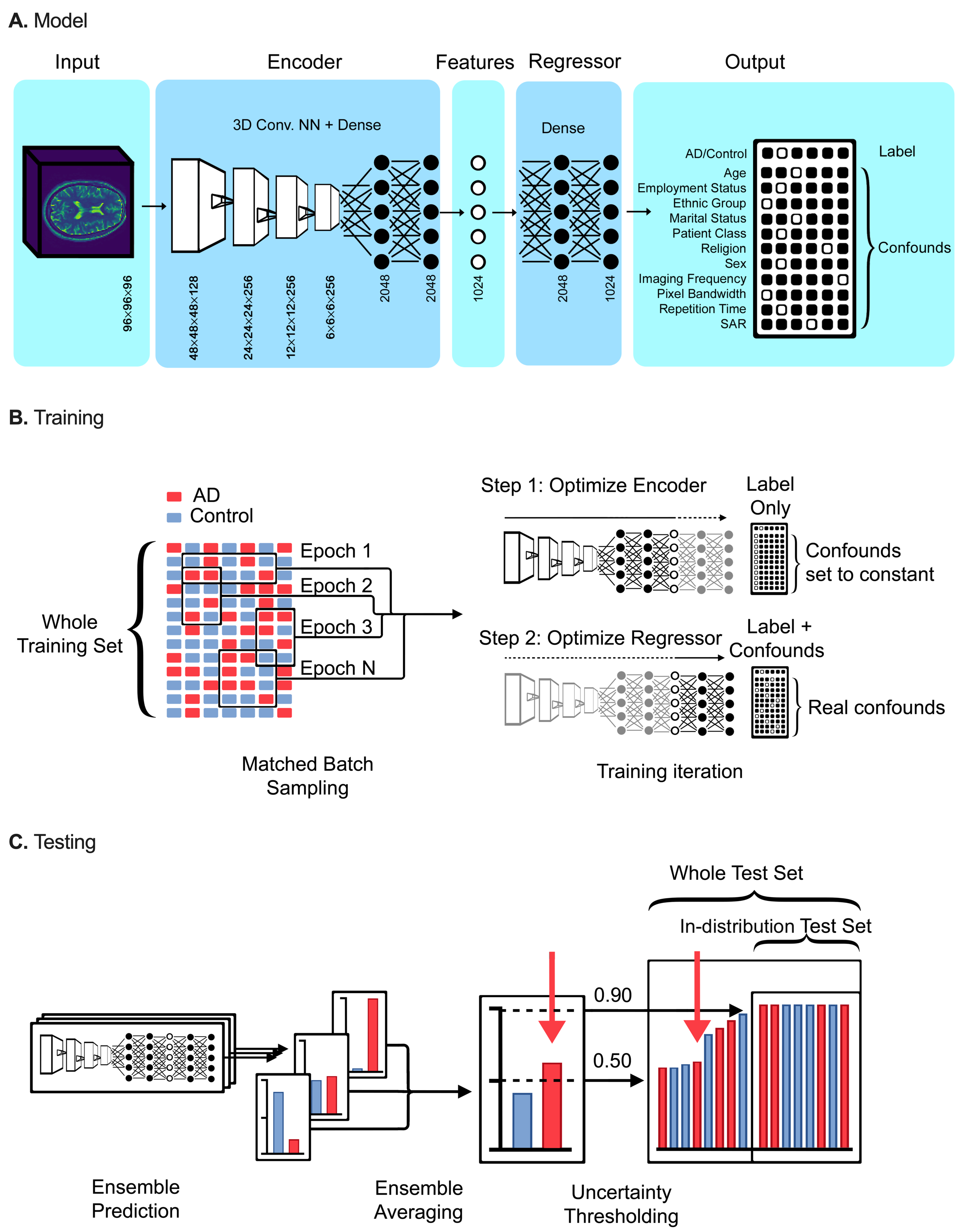}
\caption{{\bf Multi-Confound Regression Adversarial Network (MUCRAN).}
A. MUCRAN is a convolutional neural network (CNN) that takes a $96\times 96\times 96$ MRI as input, encodes it to an array of 1024 intermediary features via a CNN and a dense neural network, and regresses these features to an output array. The output array consists of one-hot binary vectors that encode both the primary label (AD/Control) and included confounding factors (sex, age, and so on). B. For training the regressed model, large batches are sampled from an imbalanced dataset such that AD and control are present in equal proportions. For each training iteration, the model is trained in a two-step adversarial process: with the regressor frozen, the encoder is fit to an output array with the label set to its real value, but each confound row set to a constant value ($[1,0,0...0]$); the regressor is then fit to the array with both the label and its true confound values. C. For testing, labels are predicted through multiple independent models, and their votes are averaged into an ensemble vote. An uncertainty thresholding is then applied to isolate an in-distribution test set.}
\label{fig:gan_pipeline_fig}
\end{figure}

%% file: figures/dataset_description.tex
\begin{figure}[!h]
\centering
\includegraphics[width=0.90\textwidth]{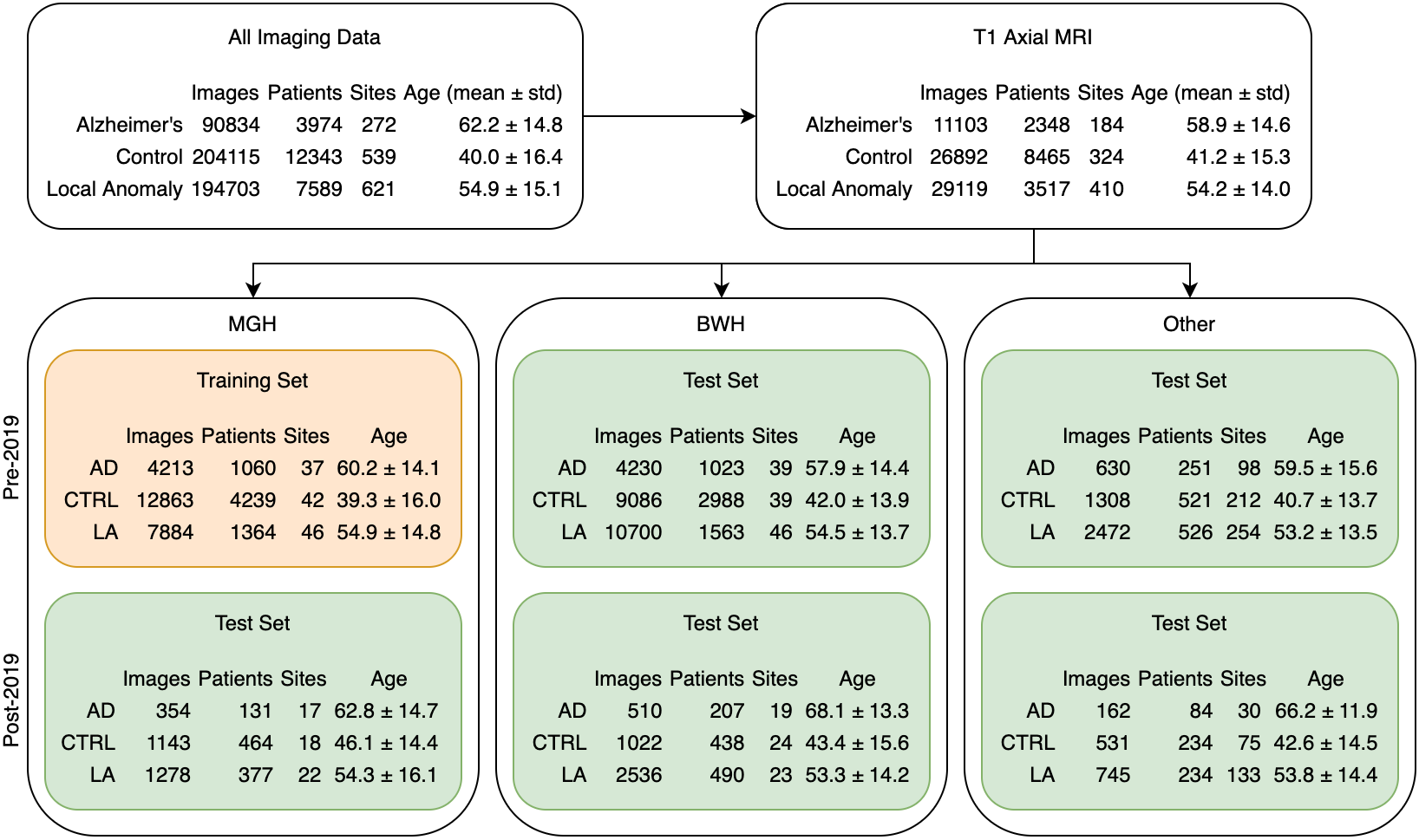}
\caption{{\bf Statistics associated with the three datasets used, pre- and post-2019.}
T1 Axial MRI, representing the plurality of structural MRI in the Mass General Brigham database, were taken from our full database of imaging data, and, from these, three test sets were isolated. "Local anomaly" refers to patients with lesions, head trauma, or tumors. The average age and standard deviation of each group is shown as well.}
\label{fig:dataset_description}
\end{figure}

%% file: figures/confounds_diagram.tex
\begin{figure}[!h]
\centering
\includegraphics[width=0.8\textwidth]{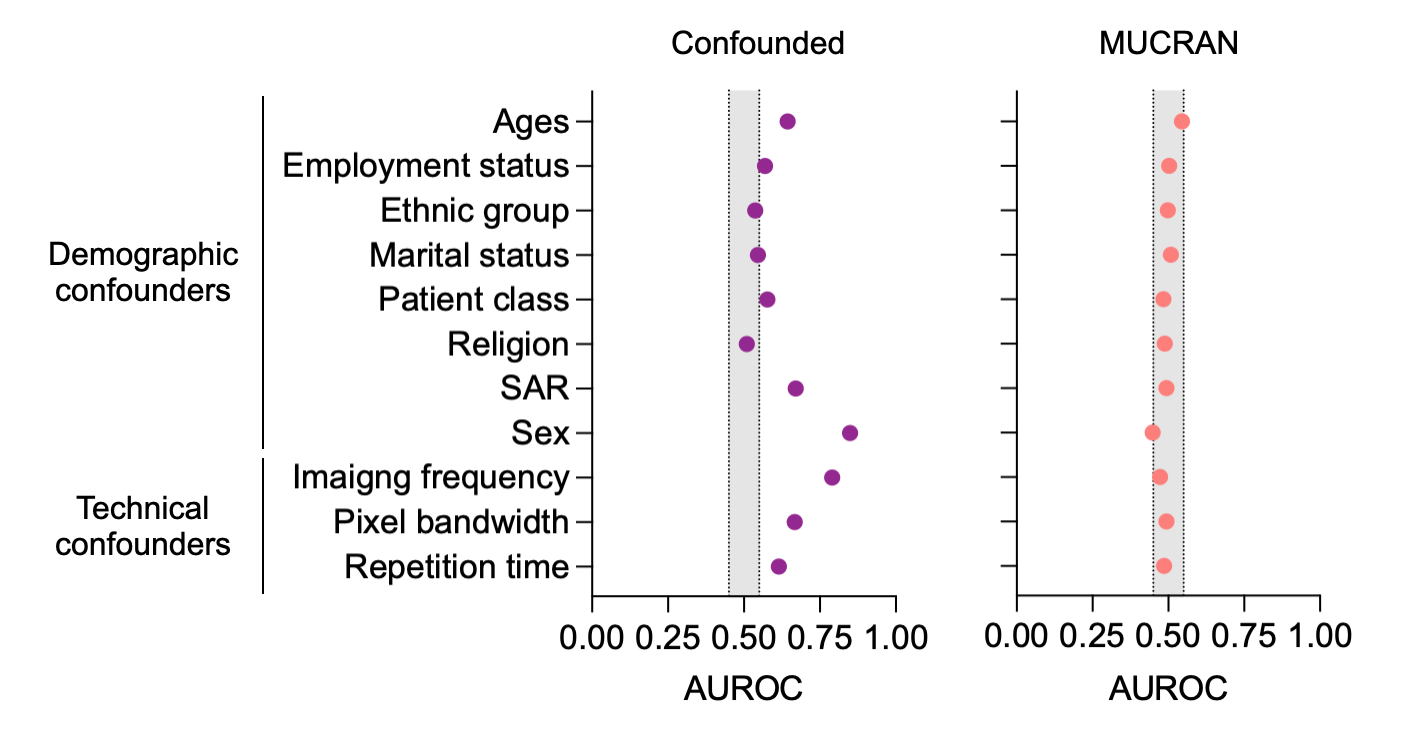}

\caption{{\bf Prediction of confounding factors.}
Averaged results of model performance for predicting demographic and technical confounding factors were reported as area under the receiver operating characteristics (AUROC) for the confounded and regressed ensembles.}
\label{fig:confounds_diagram}
\end{figure}

%% file: figures/ad_age_acc_comp.tex
\begin{figure}[!h]
\centering
{\sffamily
\centering
\includegraphics[width=0.8\textwidth]{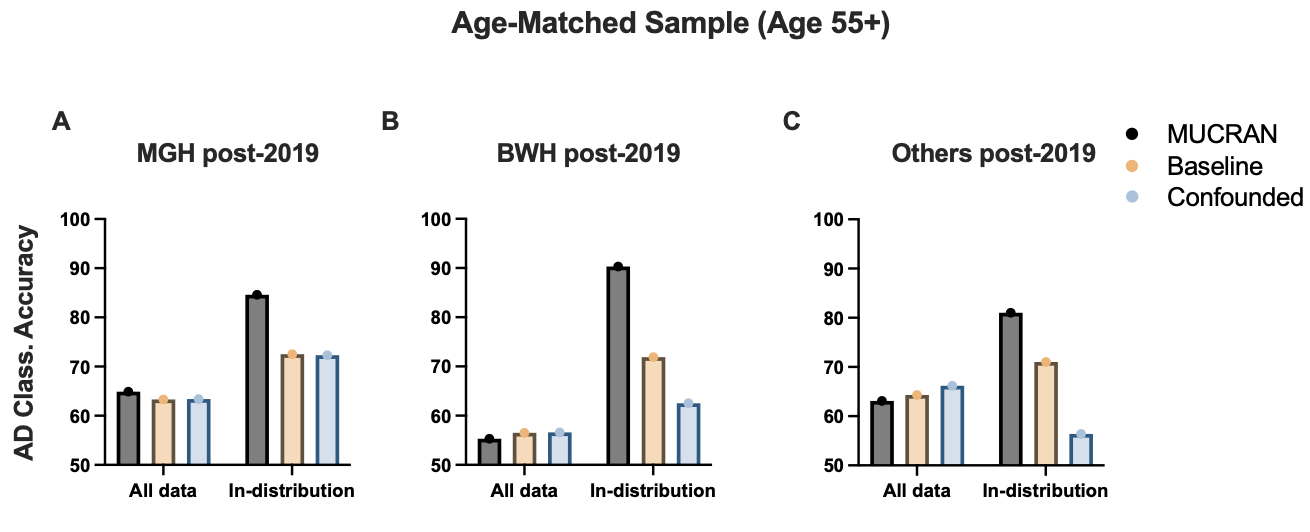}
}
\caption{{\bf Comparison of classification accuracies on post-2019 data for an age-matched sample of patients aged 55 and older in MGH, BWH, and outside hospital datasets.}
A. Results in the internal, post-2019 MGH test set. B. Results on post-2019 data from Brigham and Women's Hospital. C. Results from all outside hospital systems imported into Mass General Brigham. See Table \ref{tab:results_alz} for the full results.}
\label{fig:ad_age_acc_comp}
\end{figure}

%% file: tables/results_alz_55.tex
\begin{table}[!h]
\caption{Results on Alzheimer's classification when the whole test set is included (threshold = 0.5), and results, on five test sets and the combined test sets, for which the averaged prediction in its respective category is above 0.90 (i.e., each model in the ensemble unanimously agrees). Shown are the accuracies on the portion of the dataset included after thresholding, and the size of that test set (in parentheses). Included as well are age-matched samples. See also Figure \ref{fig:gan_pipeline_fig}C.}

\label{tab:results_alz}
\sffamily
\centering
\begin{adjustbox}{width=0.90\textwidth,totalheight=0.80\textheight,keepaspectratio}
\begin{tabular}{p{2cm}p{2cm}|l|l|lllll|l}
\multicolumn{2}{l|}{\multirow{2}{*}{Classification task}} & \multirow{2}{*}{\parbox{2cm}{Uncertainty Threshold}} & \multirow{2}{*}{Model} & \multicolumn{1}{l|}{MGH} & \multicolumn{2}{l|}{BWH} & \multicolumn{2}{l|}{Others} & \multirow{2}{*}{All} \\
\multicolumn{2}{l|}{} & & & \multicolumn{1}{l|}{Post-2019} & \multicolumn{1}{l|}{Pre-2019} & \multicolumn{1}{l|}{Post-2019} & \multicolumn{1}{l|}{Pre-2019} & Post-2019 & \\ \hline
\multicolumn{1}{l|}{\multirow{12}{*}{Alzheimer’s}} & \multirow{6}{*}{\parbox{2cm}{Age-matched sample (age 55+)}} & \multirow{3}{*}{0.90 (In-distribution)} & MUCRAN & \textbf{84.6\% (117)} & \textbf{75.4\% (471)} & \textbf{90.3\% (31)} & \textbf{90.9\% (66)} & \textbf{81.0\% (21)} & \textbf{75.3\% (857)} \\
\multicolumn{1}{l|}{} &  &  & Baseline & 72.5\% (160) & 65.4\% (771) & 71.9\% (64) & 75.7\% (181) & 71.0\% (62) & 67.4\% (1317) \\
\multicolumn{1}{l|}{} &  &  & Confounded & 72.3\% (166) & 59.6\% (971) & 62.5\% (104) & 66.5\% (179) & 56.4\% (55) & 61.8\% (1655) \\ \cline{3-10} 
\multicolumn{1}{l|}{} &  & \multirow{3}{*}{0.50 (All data)} & MUCRAN & \textbf{64.9\% (276)} & 57.0\% (2923) & 55.3\% (293) & \textbf{60.3\% (315)} & 63.1\% (141) & \textbf{58.1\% (4776)} \\
\multicolumn{1}{l|}{} &  &  & Baseline & 63.3\% (275) & \textbf{57.5\% (2923)} & 56.5\% (299) & 55.3\% (311) & 64.3\% (140) & 57.9\% (4699) \\
\multicolumn{1}{l|}{} &  &  & Confounded & 63.4\% (276) & 55.9\% (2929) & \textbf{56.6\% (309)} & 55.4\% (316) & \textbf{66.2\% (142)} & 57.2\% (4698) \\ \cline{2-10} 
\multicolumn{1}{l|}{} & \multirow{6}{*}{All} & \multirow{3}{*}{0.90 (In-distribution)} & MUCRAN & \textbf{92.5\% (548)} & \textbf{89.2\% (3721)} & \textbf{93.5\% (367)} & \textbf{92.1\% (643)} & \textbf{90.5\% (285)} & \textbf{89.1\% (5783)} \\
\multicolumn{1}{l|}{} & & & Baseline & 87.7\% (778) & 84.4\% (6198) & 89.8\% (637) & 88.4\% (976) & 91.0\% (390) & 84.7\% (9469) \\
\multicolumn{1}{l|}{} & & & Confounded & 89.5\% (829) & 84.1\% (7265) & 87.6\% (740) & 86.3\% (1138) & 90.5\% (419) & 83.9\% (10951) \\ \cline{3-10} 
\multicolumn{1}{l|}{} & & \multirow{3}{*}{0.50 (All data)} & MUCRAN & 75.8\% (1463) & 72.4\% (13303) & 75.2\% (1531) & \textbf{76.9\% (1936)} & 80.2\% (681) & 72.1\% (20857) \\
\multicolumn{1}{l|}{} & & & Baseline & 77.4\% (1463) & \textbf{72.9\% (13303)} & \textbf{76.3\% (1531)} & 74.7\% (1936) & \textbf{81.6\% (681)} & \textbf{72.4\% (20857)} \\
\multicolumn{1}{l|}{} & & & Confounded & \textbf{78.4\% (1463)} & \textbf{72.9\% (13303)} & 74.7\% (1531) & 74.7\% (1936) & 81.2\% (681) & 72.3\% (20857) \\ \hline
\end{tabular}
\end{adjustbox}
\end{table}

%% file: tables/results_sex.tex
\begin{table}[!h]
\caption{Results for sex classification}

\label{tab:results_sex}
\sffamily
\centering
\begin{adjustbox}{width=0.90\textwidth,totalheight=0.80\textheight,keepaspectratio}
\begin{tabular}{p{2cm}p{2cm}|l|l|lllll|l}
\multicolumn{2}{l|}{\multirow{2}{*}{Classification task}} & \multirow{2}{*}{\parbox{2cm}{Uncertainty Threshold}} & \multirow{2}{*}{Model} & \multicolumn{1}{l|}{MGH} & \multicolumn{2}{l|}{BWH} & \multicolumn{2}{l|}{Others} & \multirow{2}{*}{All} \\
\multicolumn{2}{l|}{} & & & \multicolumn{1}{l|}{Post-2019} & \multicolumn{1}{l|}{Pre-2019} & \multicolumn{1}{l|}{Post-2019} & \multicolumn{1}{l|}{Pre-2019} & Post-2019 & \\ \hline
\multicolumn{2}{l|}{\multirow{6}{*}{Sex}} & \multirow{3}{*}{0.90 (In-distribution)} & MUCRAN & 96.0\% (1293) & \textbf{96.8\% (11834)} & \textbf{96.3\% (1789)} & \textbf{96.8\% (2345)} & \textbf{97.2\% (723)} & \textbf{96.1\% (18836)} \\
\multicolumn{2}{l|}{} & & Baseline & 96.1\% (1279) & 94.6\% (11619) & 92.7\% (1716) & 95.5\% (2277) & 95.8\% (689) & 93.7\% (18422) \\
\multicolumn{2}{l|}{} & & Confounded & \textbf{96.5\% (1304)} & 96.0\% (11380) & 95.5\% (1782) & 96.5\% (2335) & 97.2\% (712) & 95.5\% (18327) \\ \cline{3-10} 
\multicolumn{2}{l|}{} & \multirow{3}{*}{0.50 (All data)} & MUCRAN & \textbf{85.6\% (2789)} & \textbf{85.7\% (24022)} & \textbf{83.9\% (4083)} & 87.2\% (4427) & \textbf{87.4\% (1444)} & \textbf{85.7\% (36765)} \\
\multicolumn{2}{l|}{} & & Baseline & 85.1\% (2789) & 80.1\% (24022) & 79.4\% (4083) & 84.7\% (4427) & 83.4\% (1444) & 81.7\% (36765) \\
\multicolumn{2}{l|}{} & & Confounded & 85.5\% (2789) & 85.2\% (24022) & 83.1\% (4083) & \textbf{87.4\% (4427)} & 86.6\% (1444) & 85.3\% (36765) \\ \hline
\end{tabular}
\end{adjustbox}
\end{table}

%% file: tables/methods_comparison_table.tex
\begin{table}[!h]
\sffamily
\caption{Comparison of MUCRAN with other generalizable variable regression methods used in neuroimaging}
\begin{adjustbox}{width=0.80\textwidth,totalheight=0.80\textheight,keepaspectratio}

\begin{tabular}{p{5cm}|lllp{5cm}p{2cm}}
Regression Method & Multivariate? & Hand-crafted features? & Affects training set size & Notes & References \\ \hline
Confounder-Free DL Model & No & No & No & In early tests, attempts to apply multivariate regression led to mode collapse. & \cite{Zhao2020} \\
Data Matching & Yes & No & Yes & & \cite{Leming2022}  \\
Post-Hoc Counterbalancing & No & No & Yes & & \cite{Snoek2019}  \\
Regression of extracted features & Yes & Yes & No & Refers to a class of methods, such as ComBat and Linear Mixed-Effects Models, that require significant feature reduction/extraction and treat features as independent scalar components, as images are too high-dimensional for these methods to work on them directly. The exact method of feature reduction varies between studies. & \cite{Johnson2007,Bolker2009,Todd2013,Kostro2014,Dubois2018,EspinPerez2018,Yu2018,Snoek2019,More2021} \\
\hline
MUCRAN & Yes & No & No & & Current Work \\
\end{tabular}
\end{adjustbox}
\label{tab:methods_comparison_table}
\end{table}

%% file: tables/sites.tex
\begin{table}[h!]
\caption{Sites for the training set (MGH Pre-2019), as well as the numbers of AD and Controls scanned at each}

\label{tab:sites}
\sffamily
\centering
\begin{adjustbox}{width=0.90\textwidth,totalheight=0.80\textheight,keepaspectratio}
\begin{tabular}{l|rrrrr}
MRI Site Name  & \multicolumn{1}{l}{No. AD} & \multicolumn{1}{l}{No. CTRL} & \multicolumn{1}{l}{Norm AD (No. AD / Total AD)} & \multicolumn{1}{l}{Norm CTRL (No. CTRL / Total CTRL)} & \multicolumn{1}{l}{Ratio (Norm AD : Norm CTRL)} \\ \hline
MR1WA          & 275                        & 370                          & 0.0653                                          & 0.0293                                                & 2.2243                                          \\
ERMR           & 79                         & 294                          & 0.0188                                          & 0.0233                                                & 0.8041                                          \\
LI\_MR         & 135                        & 779                          & 0.0320                                          & 0.0618                                                & 0.5186                                          \\
LI             & 958                        & 5468                         & 0.2274                                          & 0.4337                                                & 0.5243                                          \\
MR3\_ELL2      & 12                         & 83                           & 0.0028                                          & 0.0066                                                & 0.4327                                          \\
MR1CH          & 196                        & 245                          & 0.0465                                          & 0.0194                                                & 2.3941                                          \\
MR3EL2         & 171                        & 472                          & 0.0406                                          & 0.0374                                                & 1.0842                                          \\
MR1W1          & 246                        & 566                          & 0.0584                                          & 0.0449                                                & 1.3007                                          \\
PR62ELL2       & 34                         & 129                          & 0.0081                                          & 0.0102                                                & 0.7888                                          \\
MRC40168       & 228                        & 294                          & 0.0541                                          & 0.0233                                                & 2.3208                                          \\
MR1A2          & 142                        & 273                          & 0.0337                                          & 0.0217                                                & 1.5566                                          \\
PR67ELL2       & 52                         & 162                          & 0.0123                                          & 0.0128                                                & 0.9606                                          \\
MRFND1         & 228                        & 715                          & 0.0541                                          & 0.0567                                                & 0.9543                                          \\
MR1L6          & 196                        & 98                           & 0.0465                                          & 0.0078                                                & 5.9853                                          \\
MRS1OW         & 74                         & 286                          & 0.0176                                          & 0.0227                                                & 0.7743                                          \\
MR2WA          & 89                         & 98                           & 0.0211                                          & 0.0078                                                & 2.7178                                          \\
MR2EL2         & 94                         & 238                          & 0.0223                                          & 0.0189                                                & 1.1820                                          \\
EXHDMRI        & 21                         & 27                           & 0.0050                                          & 0.0021                                                & 2.3276                                          \\
MR1Y6          & 124                        & 179                          & 0.0294                                          & 0.0142                                                & 2.0731                                          \\
MR2CH          & 149                        & 187                          & 0.0354                                          & 0.0148                                                & 2.3845                                          \\
PR66ELL2       & 36                         & 252                          & 0.0085                                          & 0.0200                                                & 0.4275                                          \\
MR1NS          & 161                        & 192                          & 0.0382                                          & 0.0152                                                & 2.5095                                          \\
PR248          & 44                         & 181                          & 0.0104                                          & 0.0144                                                & 0.7275                                          \\
MRC25157       & 44                         & 64                           & 0.0104                                          & 0.0051                                                & 2.0574                                          \\
MR2Y6          & 160                        & 284                          & 0.0380                                          & 0.0225                                                & 1.6860                                          \\
MR1NY          & 2                          & 12                           & 0.0005                                          & 0.0010                                                & 0.4988                                          \\
MR1EL2         & 131                        & 275                          & 0.0311                                          & 0.0218                                                & 1.4256                                          \\
CHCMRI         & 57                         & 156                          & 0.0135                                          & 0.0124                                                & 1.0935                                          \\
MR3Y6          & 1                          & 3                            & 0.0002                                          & 0.0002                                                & 0.9975                                          \\
MRC20597       & 11                         & 52                           & 0.0026                                          & 0.0041                                                & 0.6331                                          \\
MRC35022       & 28                         & 95                           & 0.0066                                          & 0.0075                                                & 0.8820                                          \\
CHCMR2         & 4                          & 15                           & 0.0009                                          & 0.0012                                                & 0.7980                                          \\
GEMS           & 0                          & 5                            & 0.0000                                          & 0.0004                                                & 0.0000                                          \\
MEDPC          & 9                          & 12                           & 0.0021                                          & 0.0010                                                & 2.2445                                          \\
M247OC0        & 0                          & 1                            & 0.0000                                          & 0.0001                                                & 0.0000                                          \\
BAY4OC         & 2                          & 2                            & 0.0005                                          & 0.0002                                                & 2.9926                                          \\
BAY1OW0        & 0                          & 1                            & 0.0000                                          & 0.0001                                                & 0.0000                                          \\
MR3CH          & 11                         & 16                           & 0.0026                                          & 0.0013                                                & 2.0574                                          \\
BAY3OC         & 0                          & 8                            & 0.0000                                          & 0.0006                                                & 0.0000                                          \\
MR3WA          & 8                          & 16                           & 0.0019                                          & 0.0013                                                & 1.4963                                          \\
MRS3OC0        & 1                          & 2                            & 0.0002                                          & 0.0002                                                & 1.4963                                          \\
MRS5MRS5       & 0                          & 1                            & 0.0000                                          & 0.0001                                                & 0.0000                                          \\
\textbf{TOTAL} & \textbf{4213}              & \textbf{12608}               & \textbf{1.0000}                                 & \textbf{1.0000}                                       & \textbf{}                                      
\end{tabular}
\end{adjustbox}
\end{table}